\documentclass[conference, 10pt]{IEEEtran}
\IEEEoverridecommandlockouts
\usepackage{cite}
\usepackage{amsmath,amssymb,amsfonts}
\usepackage{algorithmic}
\usepackage[normalem]{ulem}
\usepackage{caption}
\usepackage{graphicx}
\usepackage{textcomp}
\usepackage{xcolor}
\usepackage{url}
\usepackage{bbm}
\usepackage{cuted}
\usepackage{algorithmic}
\usepackage{algorithm}

\newcommand{\removelatexerror}{\let\@latex@error\@gobble}
\def\BibTeX{{\rm B\kern-.05em{\sc i\kern-.025em b}\kern-.08em
    T\kern-.1667em\lower.7ex\hbox{E}\kern-.125emX}}

\begin{document}

\title{Distilling Calibration via Conformalized Credal Inference\\
\thanks{The work of J. Huang was supported by the King’s College London and China Scholarship Council for their Joint Full-Scholarship (K-CSC) (grant agreement No. 202206150005). The work of O. Simeone was supported by European Union’s Horizon Europe project CENTRIC (101096379), by the Open Fellowships of the EPSRC (EP/W024101/1) and by the EPSRC project (EP/X011852/1).}}

\author{\IEEEauthorblockN{Jiayi Huang\IEEEauthorrefmark{1}, Sangwoo Park\IEEEauthorrefmark{1}, Nicola Paoletti\IEEEauthorrefmark{2}, and Osvaldo Simeone\IEEEauthorrefmark{1}}
\IEEEauthorblockA{\IEEEauthorrefmark{1}King's Communications, Learning \& Information Processing (KCLIP) lab \\ Centre for Intelligent Information Processing Systems (CIIPS), King’s College London, London, UK}
\IEEEauthorblockA{\IEEEauthorrefmark{2}Department of Informatics,  King’s College London, London, UK  \\
Email: \{jiayi.3.huang, sangwoo.park, nicola.paoletti, osvaldo.simeone\}@kcl.ac.uk}\vspace{-10cm}}


\maketitle

\begin{strip}
    \noindent 
    \makebox[\textwidth]{\includegraphics[width=\textwidth]{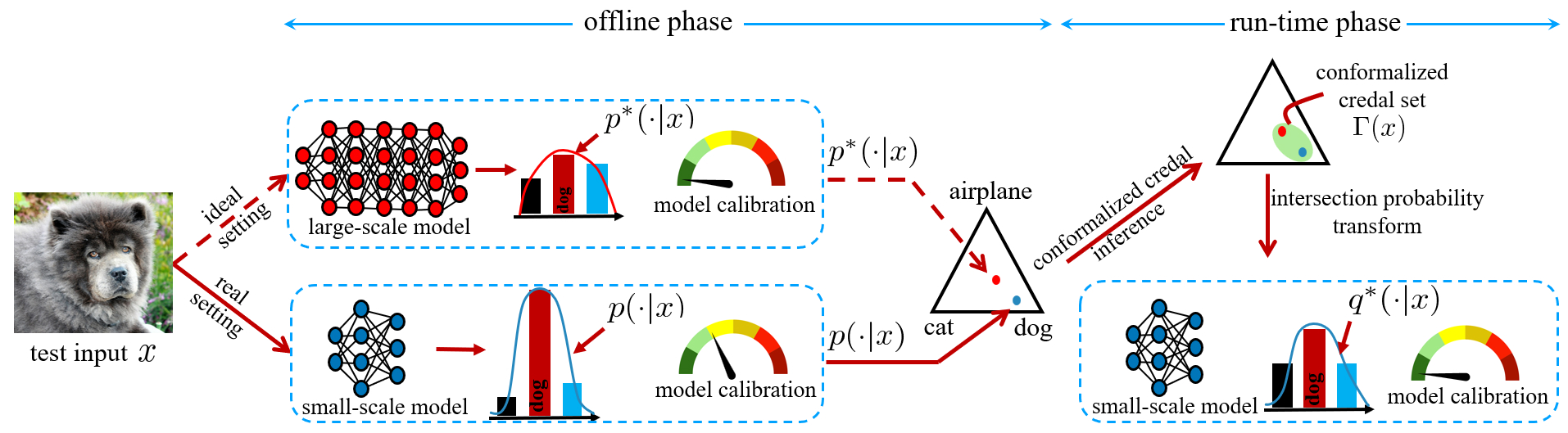}}
    \captionof{figure}{Given an input $x$, the predictive distribution ideally coincides with that of a large-scale cloud-based model $p^*(\cdot|x)$. In the setting studied in this work, a small-scale edge-based model produces a probabilistic distribution $p(\cdot|x)$ that deviates from the reference distribution $p^*(\cdot|x)$, and is thus uncalibrated. The proposed conformalized credal inference-based scheme post-processes the small-scale edge model output $p(\cdot|x)$ via a simple thresholding mechanism to produce a subset $\Gamma(x)$ in the simplex of predictive distributions, with the guarantee of containing the reference distribution $p^*(\cdot|x)$ with probability $1-\epsilon$. A final calibrated predictive distribution can be obtained via ensembling or via other combining mechanisms.}
\end{strip}

\newcommand{\np}[1]{{\color{blue}[NP:#1]}}

\begin{abstract}
Deploying artificial intelligence (AI) models on edge devices involves a delicate balance between meeting stringent complexity constraints, such as limited memory and energy resources, and ensuring reliable performance in sensitive decision-making tasks. One way to enhance reliability is through uncertainty quantification via Bayesian inference. This approach, however, typically necessitates maintaining and running multiple models in an ensemble, which may exceed the computational limits of edge devices. This paper introduces a low-complexity methodology to address this challenge by distilling calibration information from a more complex model. In an offline phase, predictive probabilities generated by a high-complexity cloud-based model are leveraged to determine a threshold based on the typical divergence between the cloud and edge models. At run time, this threshold is used to construct \emph{credal sets} -- ranges of predictive probabilities that are guaranteed, with a user-selected confidence level, to include the predictions of the cloud model. The credal sets are obtained through thresholding of a divergence measure in the simplex of predictive probabilities. Experiments on visual and language tasks demonstrate that the proposed approach, termed Conformalized Distillation for Credal Inference (CD-CI), significantly improves calibration performance compared to low-complexity Bayesian methods, such as Laplace approximation, making it a practical and efficient solution for edge AI deployments.
\end{abstract}

\begin{IEEEkeywords}
Calibration, distillation, conformal prediction, credal inference
\end{IEEEkeywords}

\section{Introduction} \label{sec:intro}

\subsection{Context and Motivation}

Modern artificial intelligence (AI) models, including neural networks and large language models, have achieved remarkable success in decision-making and generative tasks. However, two significant challenges persist in their deployment: (\emph{i}) achieving \emph{reliability} in safety-critical applications \cite{guo2017calibration}, and (\emph{ii}) enabling \emph{efficient} deployment on edge devices with constrained resources \cite{singh2023edge}. 

Traditional Bayesian methods, which model epistemic uncertainty by treating model parameters as random variables, are a popular approach to improving reliability by enhancing \emph{calibration} \cite{mackay2003information, li2024arithmetic, khan2021bayesian}. A well-calibrated model is one whose confidence levels match the true accuracy levels, thus providing trustworthy information about the reliability of the model's output.  However, Bayesian methods have several drawbacks. For one, they typically require maintaining and running multiple models in order to carry out ensembling, which is computationally challenging for edge devices. Furthermore, they depend on the choice of prior distributions, whose misspecification may lead to suboptimal calibration.

Against this background, this paper introduces a low-complexity, practical approach for calibrating edge models. Our approach distils calibration knowledge from complex cloud models, ensuring reliable performance without the computational overhead of Bayesian methods.

\subsection{Conformalized Distillation for Credal Inference}

As illustrated in Fig. 1, this paper introduces a low-complexity methodology to enhance the calibration of a small-scale edge model by distilling calibration information from a more complex cloud model. In the setting under study, prior to deployment at the edge, a small-scale model is calibrated with the use of data collected from a large-scale model. 

The approach, termed \emph{Conformalized Distillation for Credal Inference} (CD-CI),  builds on \emph{conformal prediction} (CP)  \cite{angelopoulos2024theoretical, shafer2008tutorial} and \emph{imprecise probability} (IP) \cite{walley1991statistical, beer2013imprecise}. In an \emph{offline calibration phase}, CD-CI uses the predictive probabilities generated by a high-complexity cloud-based model to determine a threshold based on the typical divergence between the predictions of the cloud and edge models. 

At \emph{run time}, CD-CI uses this threshold to construct \emph{credal sets} \cite{caprio2024Bayesian} -- ranges of predictive probabilities that are guaranteed, with a user-selected confidence level, to include the predictions of the cloud model. The credal sets are obtained through simple thresholding of a divergence measure in the simplex of predictive probabilities \cite{javanmardi2024conformalized}. Credal sets are finally converted into predictive distributions by using methods such as entropy maximization \cite{lukasiewicz2013credal} or intersection probabilities \cite{wang2024credal}. 
  
\subsection{Related Work}

\emph{Imprecise Probability} (IP) offers a mathematical framework for describing random events that cannot be captured by standard probability theory \cite{dempster2008upper, walley1991statistical, gillies2012philosophical, manski2003partial, walley1982towards, cattaneo2012likelihood, cattaneo2017empirical}. IP handles scenarios where multiple probabilities are assigned to a single event, such as: (\emph{i}) when the underlying probability distribution varies over time (e.g., the probability of rain tomorrow differs from yesterday) \cite{manski2003partial, walley1982towards}; (\emph{ii}) when estimating a fixed distribution requires considering an uncertainty set (e.g., the rain probability is estimated between $0.2$ and $0.3$) \cite{cattaneo2012likelihood}; and (\emph{iii}) when subjective beliefs, like Bayesian priors, are not uniquely defined (e.g., expecting rain 2–3 days per week in London) \cite{walley1991statistical}. For an empirical interpretation of IP, see \cite{cattaneo2017empirical}.  

\emph{Imprecise Probabilistic Machine Learning} (IPML) \cite{zaffalon2002naive, denoeux2000neural, chen2023imprecise, caprio2024Bayesian} applies IP in machine learning, addressing issues such as imprecise labelling (e.g., crowdsourced labels \cite{chen2023imprecise}), domain generalization \cite{caprio2024Bayesian}, and Bayesian model misspecification \cite{caprio2024Bayesian}. While IPML can enhance robustness and generalization \cite{chen2023imprecise, caprio2024Bayesian}, most approaches require constructing a \emph{credal set} -- a convex set of distributions containing the target distribution. This construction often relies on assumptions about the target distribution \cite{caprio2024Bayesian}, which may not hold in practice.

To address this, \emph{distribution-free IPML} \cite{cella2021valid, cella2022validity, javanmardi2024conformalized, caprio2024conformalized} uses \emph{conformal prediction} (CP) \cite{vovk2005algorithmic} to construct credal sets. Early methods \cite{cella2021valid, cella2022validity} focused on confidence intervals for labels using conformal $p$-values \cite{bates2023testing}. Recent work \cite{javanmardi2024conformalized, caprio2024conformalized} simplifies this approach by leveraging ground-truth label distributions for calibration data, justified using the concept of \emph{ambiguous} labels \cite{stutz2023conformal}. In the system proposed in this work, ground-truth label distributions are naturally provided by the cloud model.

\emph{AI Calibration} aims to adjust AI models producing probabilistic outputs, avoiding overconfidence or underconfidence~\cite{niculescu2005predicting, guo2017calibration, gupta2020calibration, mukhoti2020calibrating, bohdal2021meta, marx2024calibration} (e.g., mitigating LLM hallucinations \cite{achiam2023gpt, huang2023survey}). While Bayesian learning offers theoretical calibration \cite{simeone2022machine, zuk2012number}, it is limited in practice by model misspecification \cite{masegosa2020learning,zecchin2023robust} and computational complexity \cite{tierney1986accurate, jordan1999introduction, daxberger2021laplace}, particularly for edge devices \cite{katti2023bayesian}. Alternatives include (\emph{i}) {calibration-aware training} \cite{kumar2018trainable, lakshminarayanan2017simple, huang2024calibrating} and (\emph{ii}) {post-hoc calibration} \cite{platt1999probabilistic, vovk2012venn, marx2022modular}. Calibration-aware training modifies model training but increases complexity, especially with ensembling. Post-hoc calibration adjusts pre-trained models using a held-out dataset, though performance can degrade with limited calibration data \cite{shen2024thermometer}. CP-based post-hoc methods \cite{vovk2012venn, marx2022modular, vovk2017nonparametric} address this by producing calibrated distributions, referred to as conformal calibration or Venn predictors. This paper aims at producing calibrated credal sets, offering a more nuanced and comprehensive characterization of uncertainty compared to calibrated distributions.

Finally, this work aligns with \emph{knowledge distillation} \cite{hinton2015distilling}, where a larger model aids a smaller model. Unlike traditional approaches focused on accuracy \cite{zhu2023rethinking}, we enhance the calibration of the smaller model without compromising its accuracy.

\subsection{Main Contributions}

This paper proposes a novel, low-complexity calibration method specifically designed for edge AI deployment. The primary contributions are as follows:
\begin{itemize}
    \item \emph{Credal set construction via distilled calibration}: We introduce a post-processing approach, CD-CI, that distills calibration knowledge from large-scale cloud models. By treating the predictive distributions from cloud models as a reference, we enable edge models to make calibrated predictions with statistical reliability guarantees.
    \item \emph{Robust predictive distribution}: When a single predictive distribution is required, CD-CI extracts a distribution from the credal set by using the intersection probability approach \cite{wang2024credal}, achieving a more robust performance compared to low-complexity Bayesian methods. 
    \item \emph{Experimental validation}: We present experiments on visual and language modelling tasks, including the CIFAR-10 dataset \cite{krizhevsky2010cifar} and the SNLI dataset \cite{bowman2015large}, in which small-scale models are obtained via smaller architectures or quantized weights \cite{li2024evaluating, leviathan2023fast}.   We demonstrate significant improvements in calibration performance, as measured by the \emph{expected calibration error} (ECE) \cite{guo2017calibration},  over the original small-scale model, with negligible drops in accuracy. Comparisons are also provided with a low-complexity Bayesian learning method, the Laplace approximation \cite{daxberger2021laplace}. These results highlight the effectiveness of our approach in real-world edge deployments. 

\end{itemize}


\subsection{Organization}
The remainder of this paper is organized as follows. Sec.~\ref{sec:problem_definition} defines the problem, and the proposed methodology for distilling calibration via conformalized credal inference is presented in Sec.~\ref{sec:our_scheme}. Sec.~\ref{sec:hard_decision} describes how to extract a predictive distribution from a credal set. Finally, Sec.~\ref{sec:experiments} illustrates the experimental setting and results and Sec.~\ref{sec:conclusion} concludes the paper.

\section{Problem Definition} \label{sec:problem_definition}
\subsection{Distilling Calibration}
As illustrated in Fig. 1, we aim at calibrating a pre-trained small-scale $K$-way classifier, $p(y|x)$, with $d$-dimensional input $x \in \mathbb{R}^d$ and label $y \in \mathcal{K} = \{0, 1, \cdots, K-1\}$, by leveraging a pre-trained large-scale model, $p^*(y|x)$, and an unlabeled data set $\mathcal{D}^\text{unl} = \{{x_{i}}\}^{|\mathcal{D}^\text{unl}|}_{i = 1}$. As an example, the large-scale model may be cloud-based, while the small-scale model may be intended for edge deployment.

Given a test input $x$, the calibration procedure  \emph{post-processes} the output of the small-scale model, given by the distribution $p(\cdot|x) = \{p(y|x)\}_{y\in \mathcal{K}}$, to produce a \emph{subset} of predictive probability distributions. This subset aims at capturing the uncertainty of the small-scale model about the predictive distribution of the large-scale model, $p^*(\cdot|x)$ for the given input $x$. This uncertainty arises due to the limited computational power of the small-scale model as compared to the large-scale model. We aim to derive a low-complexity calibration procedure in which the subset is defined by a simple thresholding mechanism.


Let $\mathcal{P}$ denote the simplex of $K$-dimensional probabilistic distributions.  At test time, as mentioned, the calibration procedure maps the small-scale model output probability $p(\cdot|x)$  into a subset $\Gamma(x) \subseteq \mathcal{P}$. The mapping between predictive probabilities and a subset of $\mathcal{P}$ is designed during an offline \emph{distillation} phase in which the designer has access to the unlabeled data set $\mathcal{D}^{\text{unl}}$ and to the large-scale model. The large-scale model is no longer accessible at test time. 

The design goal is to ensure that the set includes the reference distribution $p^*(\cdot|x)$ that would have been produced by the large-scale model with probability no smaller than a user-defined level $1-\epsilon$, i.e.,
\begin{align} \label{eq:coverage}
    \Pr \left[p^*(\cdot|x) \in  \Gamma(x) \right] \geq 1 - \epsilon,
\end{align}
where $\epsilon \in [0,1]$ is the desired \emph{miscoverage} rate. The probability in (\ref{eq:coverage}) is evaluated with respect to the distribution of the unlabeled data set $\mathcal{D}^{\text{unl}}$ used to design the post-processing mechanism, as well as over the test input $x$. 

The condition (\ref{eq:coverage}) can be satisfied for any miscoverage rate $\epsilon$ by setting $\Gamma(x) = \mathcal{P}$, i.e., by producing the set of all possible distributions on the label set $\mathcal{K}$ in response to any test input $x$. However, this output would be clearly uninformative. Therefore, to gauge the informativeness of the set predictor, we evaluate the normalized average size of the set, also known as \emph{inefficiency}, i.e.,
\begin{align} \label{eq:inefficiency}
     \mathbb{E} \left[ \frac{|\Gamma(x)|}{|\mathcal{P}|} \right].
\end{align}
The expectation in (\ref{eq:inefficiency}) is taken with respect to the same distribution as in (\ref{eq:coverage}). Furthermore, the size $|\Gamma(x)|$ corresponds to the standard volume covered by set $\Gamma(x)$ within the simplex $\mathcal{P}$.

\section{Conformalized Credal Inference} \label{sec:our_scheme}
In this section, we first introduce credal sets, and then we propose a way to conformalize the credal sets so as to satisfy the coverage requirement (\ref{eq:coverage}).

\subsection{Credal Sets} 
Given a test input $x$ and an output $p(\cdot|x)$ of the small-scale model, the set predictor $\Gamma(x)$ is constructed by including all distributions $q = q(\cdot|x)$ in a neighbourhood of $p(\cdot|x)$. The radius defining the size of the neighbourhood is determined during the offline calibration phase by using the unlabeled data and the large-scale model.

To elaborate, consider the class of divergences between two distributions $q_1$ and $q_2$ in simplex $\mathcal{P}$, i.e.,
\begin{align} \label{eq:f-diveregnce}
    \mathrm{D}_f (q_1 \| q_2) = \mathbb{E}_{z \sim q_2(z)} \left[f \left(\frac{q_1(z)}{q_2(z)}\right) \right],
\end{align}
where $f(\cdot)$ is a convex function satisfying the properties (\emph{i}) $f(1) =0$, and (\emph{ii}) $0 \cdot f(0/0) = 0$ \cite{Polyanskiy_Wu_2024, simeone_cqit}. The class of $f$-divergences encompasses a variety of divergence measures, including the Kullback–Leibler (KL) divergence and the Tsallis divergence.

Given a $f$-divergence $\mathrm{D}_f (\cdot \| \cdot)$, we obtain the credal set as 
\begin{align} \label{eq:credal_set}
    \Gamma (x) = \left \{ q \in \mathcal{P} : \hspace{0.5em} \mathrm{D}_f (q \| p(\cdot|x)) \leq \gamma \right \},
\end{align}
where $\gamma$ is a threshold to be determined during the offline calibration phase.

\subsection{Distilling Calibration via Conformalized Credal Inference}
Conformalized credal inference determines the threshold $\gamma$ in (\ref{eq:credal_set}) to guarantee the coverage condition (\ref{eq:coverage}). This is done offline by leveraging the unlabeled data set $\mathcal{D}^{\text{unl}}$. To this end, we first construct the calibration data set
\begin{align} \label{eq:calibration_dataset}
    \mathcal{D}^\text{cal} = \{{(x_{i}, p^*(\cdot|x_i))}\}^{|\mathcal{D}^\text{unl}|}_{i = 1},
\end{align}
in which the large-scale model is used to assign \emph{soft} label $p^*(\cdot|x_i)$ to the unlabeled input $x_i$. Then, we leverage the data  $\mathcal{D}^\text{cal}$ to construct the data set
\begin{align} \label{nc-score-set}
    \mathcal{S} = \{ s_i = \mathrm{D}_f(p^*(\cdot|x_i) \| p(\cdot|x_i))) \}_{i=1}^{|\mathcal{D}^{\text{cal}}|}.
\end{align} 
The data set $\mathcal{S}$ collects the divergence values $\mathrm{D}_f(p^*(\cdot|x_i) \| p(\cdot|x_i))$ between the reference and the predictive distributions in the unlabeled data set $\mathcal{D}^{\text{unl}}$. Thus, this data set supports inferences about the distributions of the divergence $\mathrm{D}_f(p^*(\cdot|x_i) \| p(\cdot|x_i))$ as the input $x$ varies according to the underlying population distribution.

Following the standard split conformal prediction (CP) methodology, we compute the $ \lceil (1 + |\mathcal{D}^{\text{cal}}|)(1 - \epsilon) \rceil$ smallest element in set $\mathcal{S}$, i.e., the $ \lceil (1 + |\mathcal{D}^{\text{cal}}|)(1 - \epsilon) \rceil/|\mathcal{D}^{\text{cal}}|$-th empirical quantile of the divergences in the data set $\mathcal{S}$. Finally, the radius threshold in (\ref{eq:credal_set}) is evaluated as 
\begin{align} \label{eq:threshold}
    \gamma = \lceil (1 + |\mathcal{D}^{\text{cal}}|)(1 - \epsilon) \rceil \text{ smallest element of } \mathcal{S}.
\end{align}

The proposed method, referred to as Conformalized Distillation for Credal Inference (CD-CI), is summarized in Algorithm~\ref{alg:CD_CI_offline} and Algorithm~\ref{alg:CD_CI_run time}. Specifically, Algorithm~\ref{alg:CD_CI_offline} defines the offline phase producing the threshold $\gamma$, while Algorithm~\ref{alg:CD_CI_run time} describes the run-time operation.


\begin{figure}[htb]   
  \renewcommand{\algorithmicrequire}{\textbf{Input:}}
  \renewcommand{\algorithmicensure}{\textbf{Output:}}
  \begin{algorithm}[H] 
    \caption{Conformalized Distillation for Credal Inference (CD-CI) -- Offline Phase}
    \begin{algorithmic}[1] 
      \REQUIRE Unlabeled data set $\mathcal{D}^{\text{unl}}$, small-scale classifier $p(\cdot|x)$, large-scale classifier $p^*(\cdot|x)$, divergence measure $\mathrm{D}_f (\cdot \| \cdot)$, target coverage level $1-\epsilon$
      \ENSURE Threshold $\gamma$
      \STATE Construct the calibration data set $\mathcal{D}^\text{cal}$ in (\ref{eq:calibration_dataset}) by using the unlabeled data set $\mathcal{D}^{\text{unl}}$ and the large-scale classifier $p^*(\cdot|x)$
      \STATE Evaluate the data set $\mathcal{S}$ in (\ref{nc-score-set}) 
      \STATE Evaluate the radius threshold $\gamma$ in (\ref{eq:threshold}), for the given target coverage level $1-\epsilon$ 
      \STATE {\textbf{return } Threshold $\gamma$}
    \end{algorithmic} \label{alg:CD_CI_offline}
  \end{algorithm}
\end{figure}

\begin{figure}[htb]   
  \renewcommand{\algorithmicrequire}{\textbf{Input:}}
  \renewcommand{\algorithmicensure}{\textbf{Output:}}
  \begin{algorithm}[H] 
    \caption{Conformalized Distillation for Credal Inference (CD-CI) -- Run-Time Operation}
    \begin{algorithmic}[1] 
      \REQUIRE Test input $x$, small-scale classifier $p(\cdot|x)$, divergence measure $\mathrm{D}_f (\cdot \| \cdot)$, simplex $\mathcal{P}$, target coverage level $1-\epsilon$
      \ENSURE Prediction set $\Gamma(x)$ or predictive distribution $q^*(\cdot|x)$
      \STATE Construct prediction set $\Gamma(x)$ using (\ref{eq:credal_set})
      \IF{predictive distribution is required}
        \STATE Derive bounds $ q_{L_y}$ and $ q_{U_y}$ as in (\ref{eq:credal_bounds})
        \STATE Compute the predictive distribution $q^*(\cdot|x)$ as in (\ref{eq:intersection_non_normalized})
      \ENDIF
      \STATE {\textbf{return } Prediction set $\Gamma(x)$ or predictive distribution $q^*(\cdot|x)$}
    \end{algorithmic} \label{alg:CD_CI_run time}
  \end{algorithm}
\end{figure}
In the offline phase, CD-CI only requires the evaluation of the threshold $\gamma$ in (\ref{eq:credal_set}). This operation requires ordering the nonconformity scores $s_i$ in the data set $\mathcal{S}$. This is a low-complexity operation with order $\mathcal{O}(|\mathcal{D}^\text{cal}|\log(|\mathcal{D}^\text{cal}|))$. In the run-time phase, as defined in Algorithm~\ref{alg:CD_CI_run time}, CD-CI evaluates the set (\ref{eq:credal_set}), from which a predictive distribution can be evaluated as detailed in the next section.


\subsection{Theoretical Reliability Guarantees}

CD-CI satisfies the following reliability guarantees.

\textbf{Theorem 1:} \textit{If the data samples in the unlabeled data set $\mathcal{D}^{\text{unl}}$ and the test input $x$ are exchangeable, e.g., i.i.d.,  then CD-CI (Algorithm~\ref{alg:CD_CI_offline} and Algorithm~\ref{alg:CD_CI_run time}) produces credal sets $\Gamma(x)$ in (\ref{eq:credal_set}) that satisfy the coverage condition
\begin{align} \label{eq:theorem_1}
    \Pr \left[p^*(\cdot|x) \in  \Gamma(x) \right] \geq 1 - \epsilon.
\end{align}
}

The proof of this theorem follows directly from the marginal coverage guarantees of CP \cite[Eq.~(1)]{angelopoulos2021gentle} (see also \cite[Thm.~4.1]{javanmardi2024conformalized}).


\section{Predictive Distributions from Credal Sets} \label{sec:hard_decision}

The proposed CD-CI method provides a low-complexity procedure to identify, for any given input $x$, a subset $\Gamma(x)$ of predictive distributions $q(\cdot|x)$ that are likely to contain the golden-standard distribution $p^*(\cdot|x)$ of the large-scale model. The set can directly provide actionable information. For example, if the set is deemed to be too large, an edge device may conclude that the local model is insufficiently accurate for the given input $x$, refraining from making a decision. 

In practice, the edge device may wish to produce a single predictive probability $q^*(\cdot|x)$ for decision-making. In this case, it is desirable that the resulting predictive probability $q^*(\cdot|x)$, evaluated from the set $\Gamma(x)$, be better calibrated than the initial distribution $p(\cdot|x)$ produced by the small-scale model. 

The distribution $q^*(\cdot|x)$ can be computed from the set $\Gamma(x)$ in different ways, which are explored in the literature on imprecise probabilities \cite{stutz2023conformal}. For example, one can choose the distribution within the credal region $\Gamma(x)$ that achieves the maximum Shannon entropy, thus finding the most conservative decisions \cite{caprio2024Bayesian}.


In this work, we adopt the \emph{intersection probability} \cite{wang2024credal} as the mechanism to obtain the final predictive distribution  $q^*(\cdot|x)$. This method is presented in Sec.~\ref{sec:intersection_prob}. Then, for reference, Sec.~\ref{sec:laplace_approx} discusses a standard low-complexity Bayesian approach that can produce a re-calibrated predictive distribution, namely the Laplace approximation \cite{daxberger2021laplace, simeone2022machine}. Finally, Sec.~\ref{sec:ECE} reviews the expected calibration error (ECE), which can be used to evaluate the calibration of predictive distributions.


\subsection{Intersection Probability} \label{sec:intersection_prob}
Given a credal set $\Gamma(x)$, the intersection probability method obtains a single predictive distribution $q^*(\cdot|x)$ as follows.

First, for every class $y \in \mathcal{K}$, one obtains lower bound $q_{L_y}$ and upper bound $q_{U_y}$ on the probability $q(y|x)$ assigned by distributions in subset $\Gamma(x)$ as
\begin{align} \label{eq:credal_bounds}
    q_{L_y} = \min_{q(\cdot|x) \in \Gamma(x)} q(y|x), \quad q_{U_y} = \max_{q(\cdot|x) \in \Gamma(x)} q(y|x).
\end{align}
In practice, the set can be represented by a discrete subset of distributions $q(\cdot|x) \in \Gamma(x)$. They can be obtained via grid search, with complexity linear in the grid size, or via importance sampling.


Finally, the predictive distribution $q^*(\cdot|x)$ is evaluated as
\begin{align} \label{eq:intersection_non_normalized}
    q^*(y|x) = q_{L_y} + \beta \cdot (q_{U_y} - q_{L_y}),
\end{align}
where $\beta \in [ 0,1 ]$ is a constant chosen to guarantee that the function $q^*(\cdot|x)$ is a valid probabilistic distribution. Accordingly, the constant value $\beta$ is calculated as 
\begin{align} \label{eq:intersection_normalized_constant}
    \beta = \frac{1 - \sum_{y=0}^{K-1}q_{L_y}}{\sum_{y=0}^{K-1} (q_{U_y} - q_{L_y})}.
\end{align}
The distribution in (\ref{eq:intersection_non_normalized}) ensures that each class $y$ is treated equally. This is done by choosing the probability $q^*(y|x)$ to be within the interval $[q_{L_y}, q_{U_y}]$ at the same relative position determined by the fraction $\beta$ for all $y \in \mathcal{K}$.

Sec.~\ref{sec:experiments} will compare the performance of the predictive distribution (\ref{eq:intersection_non_normalized}) with other methods used in the literature on imprecise probabilities, namely max entropy \cite{lukasiewicz2013credal}. A hard decision $\hat{y}$ can be obtained from the predictive distribution $q^*(\cdot|x)$ as
\begin{align} \label{eq:hard_decision}
    \hat{y}(x) = \arg \max_{y\in \mathcal{K}} q^*(y|x).
\end{align}

\subsection{Low-Complexity Bayesian Learning via the Laplace Approximation} \label{sec:laplace_approx}
Given a pre-trained small-scale model $p(\cdot|x)$, the Laplace approximation provides a low-complexity method to re-calibrate a pre-trained model using Bayesian principles \cite{mackay2003information, simeone2022machine}.
\setcounter{figure}{1}
\begin{figure*} [tb] 
    \centering
    \centerline{\includegraphics[scale=0.27]{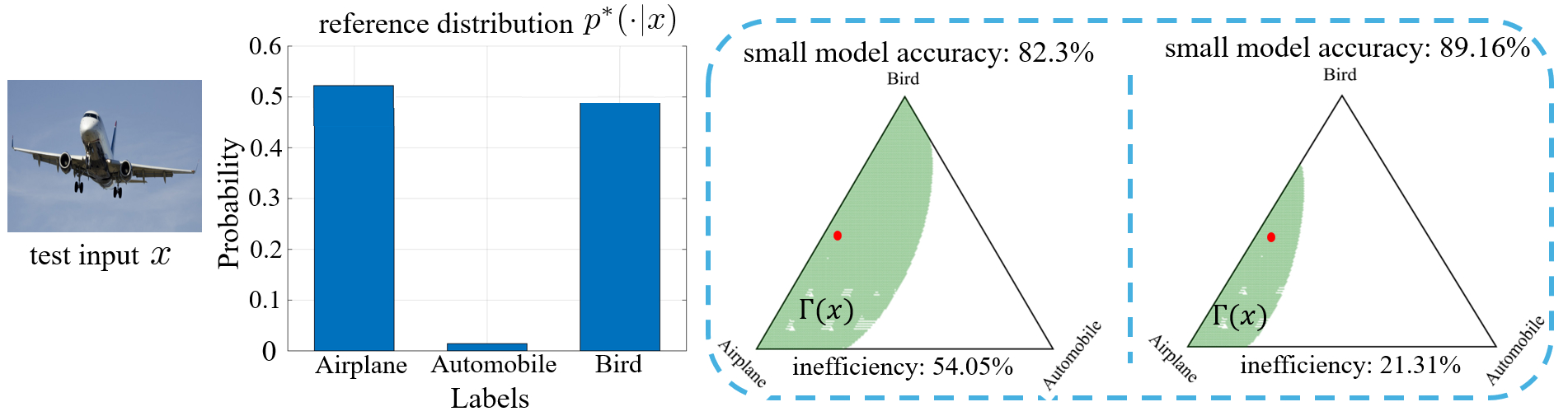}}
    \caption{Test input $x$, reference distribution $p^*(\cdot | x)$ from the large-scale model, and credal sets produced by CD-CI for small-scale models with different accuracy on the CIFAR-10 data set with classes $\{\text{airplane, automobile, bird}\}$ using the KL divergence in (\ref{eq:credal_set}) with target coverage rate $1-\epsilon = 0.9$. Note that the large-model distribution  $p^*(\cdot | x)$ is marked as red point in the simplex.} 
    \label{result:cifar_visualization} 
\end{figure*}
To elaborate, assume a parametric small-scale model
\begin{align}
    p(y|x) = p(y|x,\hat{\theta}),
\end{align}
where $\hat{\theta}$ is a pre-trained tensor of parameters. Laplace approximation approximates the posterior distribution of model parameter $\theta$ given a training data set $\mathcal{D}$ as
\begin{align} \label{eq:laplace_posterior}
    p(\theta|\mathcal{D}) \approx \mathcal{N}(\hat{\theta}, \Sigma), \quad \text{with} \quad \Sigma := \left( \nabla^2_{\theta} \mathcal{L}(\theta|\mathcal{D}) \big|_{\theta = \hat{\theta}} \right)^{-1},
\end{align}
where $\mathcal{L}(\theta|\mathcal{D})$ is the cross-entropy training loss \cite{daxberger2021laplace}. In practice, the covariance matrix $\Sigma$ can be approximated in several ways, e.g., via the Gauss-Newton method \cite{lecun1989optimal}. 

Given the approximate posterior (\ref{eq:laplace_posterior}), a predictive distribution can be made via Bayesian model ensembling as
\begin{align} \label{eq:laplace_prob}
    q^{\text{La}}(y|x) = \mathbb{E}_{\theta \sim p(\theta|\mathcal{D})} \left[ p(y|x,\theta) \right].
\end{align}
The average in (\ref{eq:laplace_prob}) can be estimated via Monte Carlo sampling from the distribution (\ref{eq:laplace_posterior}).

\subsection{Expected Calibration Error} \label{sec:ECE}
A probabilistic predictor $q(y|x)$ is said to be \emph{perfectly calibrated} if the equality
\begin{align} \label{eq:perfect_cal}
    \Pr \left[ y = \hat{y}(x) | q(\hat{y}(x)|x) = r \right] = r,
\end{align}
holds for all $r \in [0,1]$, where $\hat{y}(x) = \arg \max_{y\in \mathcal{K}} q(y|x)$. The probability in (\ref{eq:perfect_cal}) is taken over the input-output data pair $(x,y)$. By the condition (\ref{eq:perfect_cal}), the confidence level, $r = q(\hat{y}(x)|x)$ of the probabilistic predictor coincides with the probability of correct prediction for all possible $r \in [0,1]$.

The ECE quantifies the extent to which the predictor $q(y|x)$ satisfies the equality (\ref{eq:perfect_cal}). To review the ECE, we first divide the confidence interval $r \in [0,1]$ into $M$ bins $B_m = [(m-1)/M, m/M]$ for $m=1,\cdots,M$. The ECE is defined as the expected distance between the per-bin confidence $\mathrm{conf}(B_{m})$ and per-bin accuracy $\mathrm{acc}(B_{m}$) as \cite{guo2017calibration}
\begin{align} \label{eq:ECE}
    \mathrm{ECE} = \sum^{M}_{m=1} \frac{|B_{m}|}{ \sum_{m'=1}^M |B_{m'}|} \left | \mathrm{acc}(B_{m}) - \mathrm{conf}(B_m) \right |,
\end{align}
where $|B_m|$ is the number of samples whose confidence level $r$ falls in the bin $(\frac{m-1}{M}, \frac{m}{M}]$; the per-bin confidence is $\mathrm{conf}(B_{m}) = 1/|B_m|\sum_{i\in B_m}r_i$; and the per-bin accuracy is $\mathrm{acc}(B_{m}) = 1/|B_m|\sum_{i\in B_m} \mathbbm{1}(y_i=\hat{y}(x_i))$.

\section{Experimental Results} \label{sec:experiments}
In this section, we report empirical results for visual and natural language tasks. 

\subsection{Performance Metrics}
For both tasks, we consider the following evaluation metrics: (\emph{i}) \emph{inefficiency}, which evaluates the average size of the credal set $\Gamma(x)$ as in (\ref{eq:inefficiency}); (\emph{ii}) \emph{coverage}, which is the percentage of samples for which the large-scale model predictive distribution $p^*(\cdot|x)$ falls inside the credal set $\Gamma(x)$ as in (\ref{eq:coverage}); (\emph{iii}) the \emph{ECE} (\ref{eq:ECE}); and (\emph{iv}) \emph{accuracy}, measured by the probability that the hard decision obtained as in (\ref{eq:hard_decision}) is correct.

\subsection{Implementations}
Throughout, we use the $\alpha$-divergence to evaluate the conformalized credal set in (\ref{eq:credal_set}). The $\alpha$-divergence is obtained from the general definition of the $f$-divergence in (\ref{eq:f-diveregnce}) with $f(t) = (t^\alpha -1) / (\alpha(\alpha-1))$. Note that, for $\alpha = 1$, the $\alpha$-divergence reduces to the KL divergence \cite{cichocki2010families}. Increasing the parameter $\alpha$ yields more constrained sets (\ref{eq:credal_set}), for which the support of the distributions $q(\cdot|x)$ is increasingly forced not to exceed the support of the small-scale distribution $p(\cdot|x)$ \cite{minka2005divergence}.

All the experiments reported in this paper are implemented via PyTorch \cite{paszke2019pytorch} and run over a GPU server with a single NVIDIA A100 card \footnote{Code can be found at \url{https://github.com/kclip/Distilling-Calibration}.}.

\subsection{Image Classification}
For the image classification task, as in \cite{caprio2024conformalized}, we extract the first three classes from the CIFAR-10 \cite{krizhevsky2010cifar} data set. Furthermore, we adopt the ResNet-18 model \cite{he2016deep} and the Mini-VGG-8 model \cite{simonyan2014very} as the large-scale and small-scale models, respectively. 

To start, in Fig.~\ref{result:cifar_visualization}, we visualize the impact of the small-scale model accuracy on the inefficiency of the credal set produced by CD-CI via Algorithm~\ref{alg:CD_CI_offline}. We control the accuracy of the small-scale model by training over different numbers of iterations, and we set $\alpha=1$, thus relying on the KL divergence. The figure illustrates a test example $x$, the large-scale model's predictive distribution $p^*(\cdot|x)$, and credal sets produced by CD-CI for small-scale models with different accuracy on the test data set. As seen, a more accurate small-scale model yields smaller credal sets.

\begin{figure} [tb] 
    \centering
    \centerline{\includegraphics[scale=0.223]{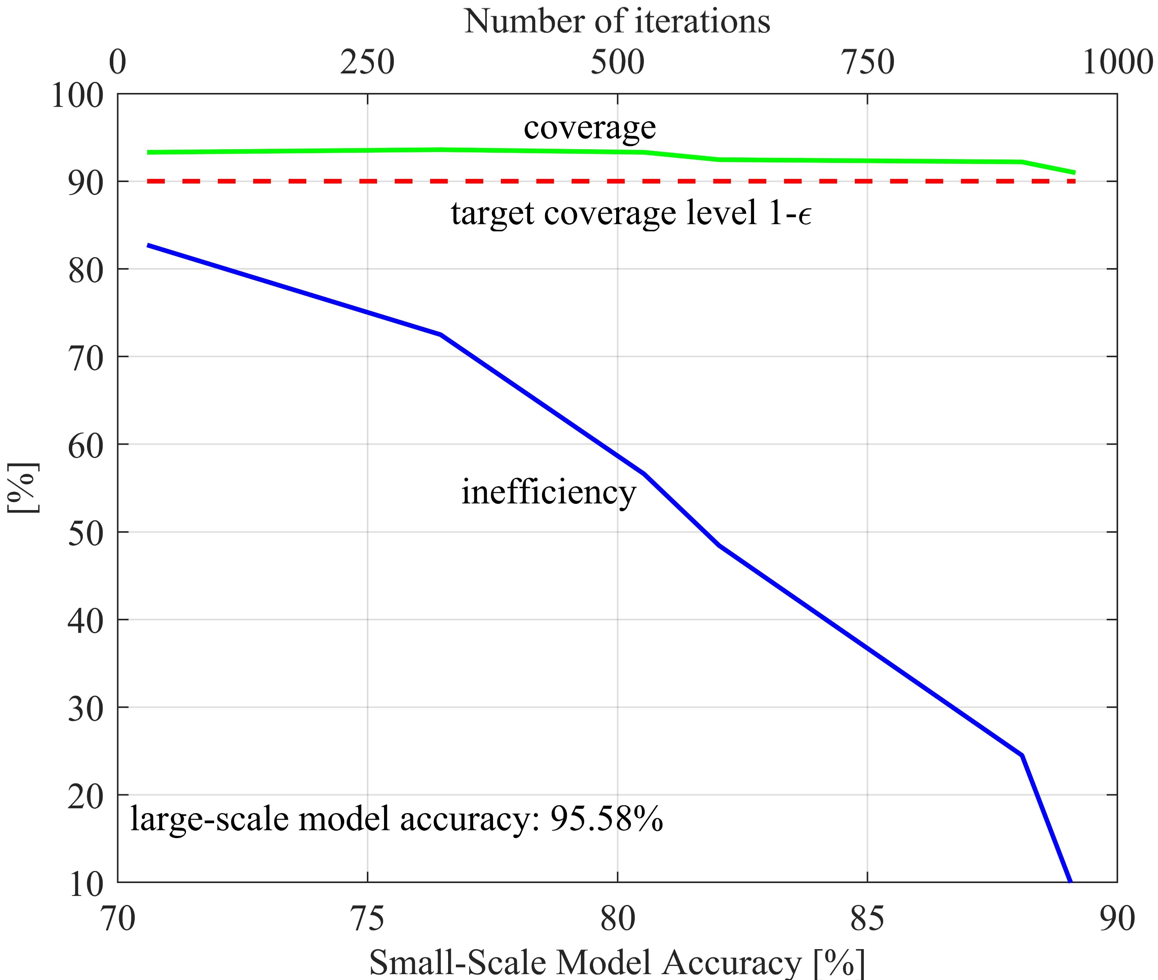}}
    \caption{Coverage and inefficiency versus the small-scale models accuracy on the CIFAR-10 data set with classes $\{\text{airplane, automobile, bird}\}$ using the KL divergence in (\ref{eq:credal_set}) with target coverage rate $1-\epsilon = 0.9$. The accuracy of the large-scale model, ResNet-18 network, is $95.58\%$, and the accuracy of the small-scale model, Mini-VGG-8, is controlled by training over different numbers of iterations. }
    \label{result:cifar_changing_accuracy} 
\end{figure}

The inefficiency and coverage of CD-CI as a function of the test accuracy of the small-scale model are shown in Fig.~\ref{result:cifar_changing_accuracy}. Note that the accuracy of the small-scale model ranges from $70\%$ to $90\%$, while the large-scale model obtains accuracy $95\%$. We set the target miscoverage rate as $\epsilon = 0.1$. Improving the accuracy of the small-scale model is seen to be instrumental in enhancing the efficiency of the conformalized credal set. Furthermore, CD-CI maintains a coverage rate close to the target level $1-\epsilon = 0.9$, for all small-scale models, validating Theorem 1.

\begin{figure} [tb] 
    \centering
    \centerline{\includegraphics[scale=0.045]{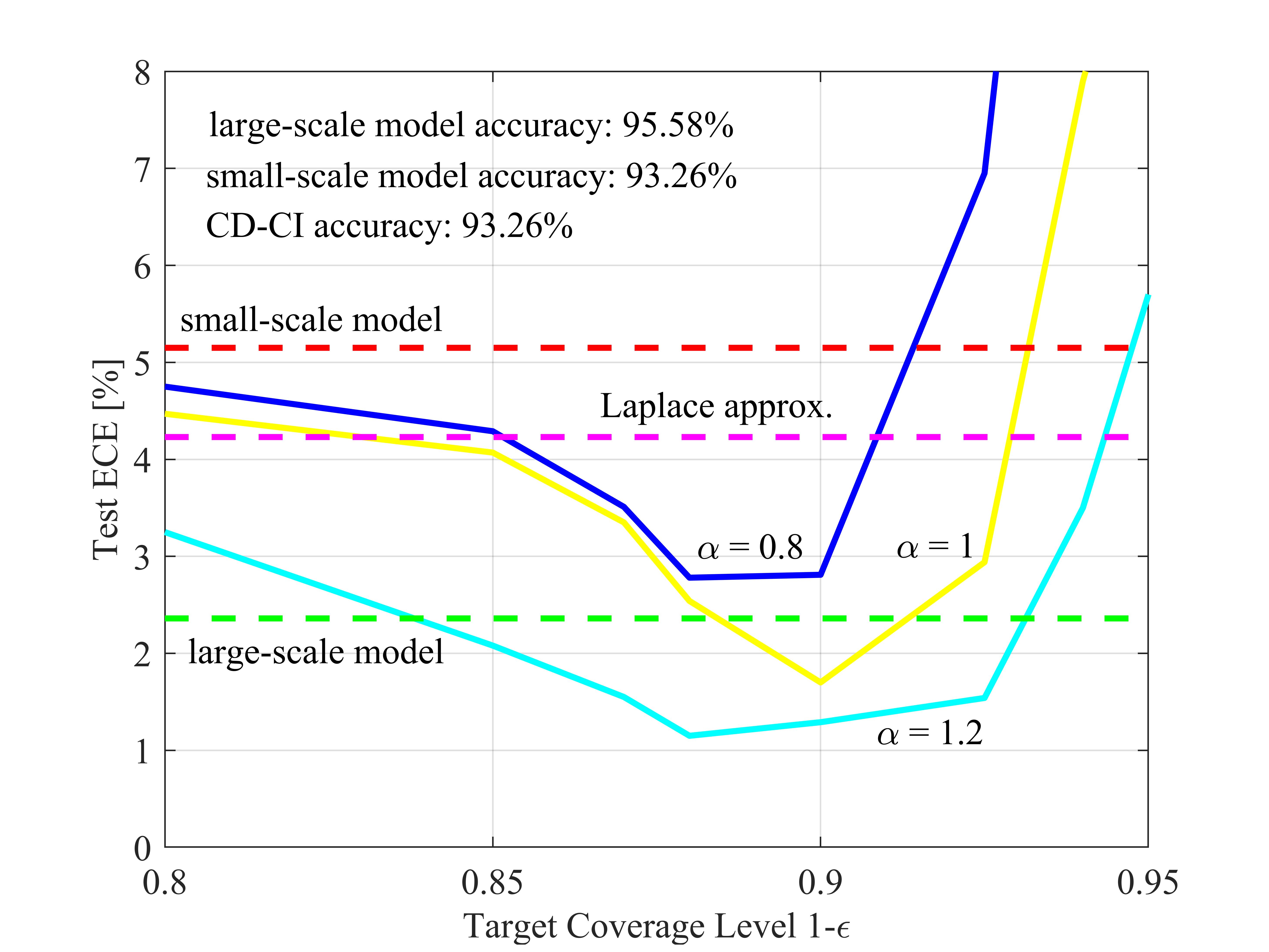}}
    \caption{ECE versus target coverage rate $1-\epsilon$ for different values of $\alpha$ for the $\alpha$-divergence used in (\ref{eq:credal_set}) on the CIFAR-10 data set with classes $\{\text{airplane, automobile, bird}\}$. The dashed lines report the ECE performance of the large-scale model predictive distribution $p^*(\cdot|x)$, of the small-scale model $p(\cdot|x)$, and of the Laplace approximation method $q^{\text{La}}(\cdot|x)$ in (\ref{eq:laplace_prob}).}
    \label{result:cifar_ECE_versus_coverage} 
\end{figure}

\begin{figure} [htb] 
    \centering
    \centerline{\includegraphics[scale=0.045]{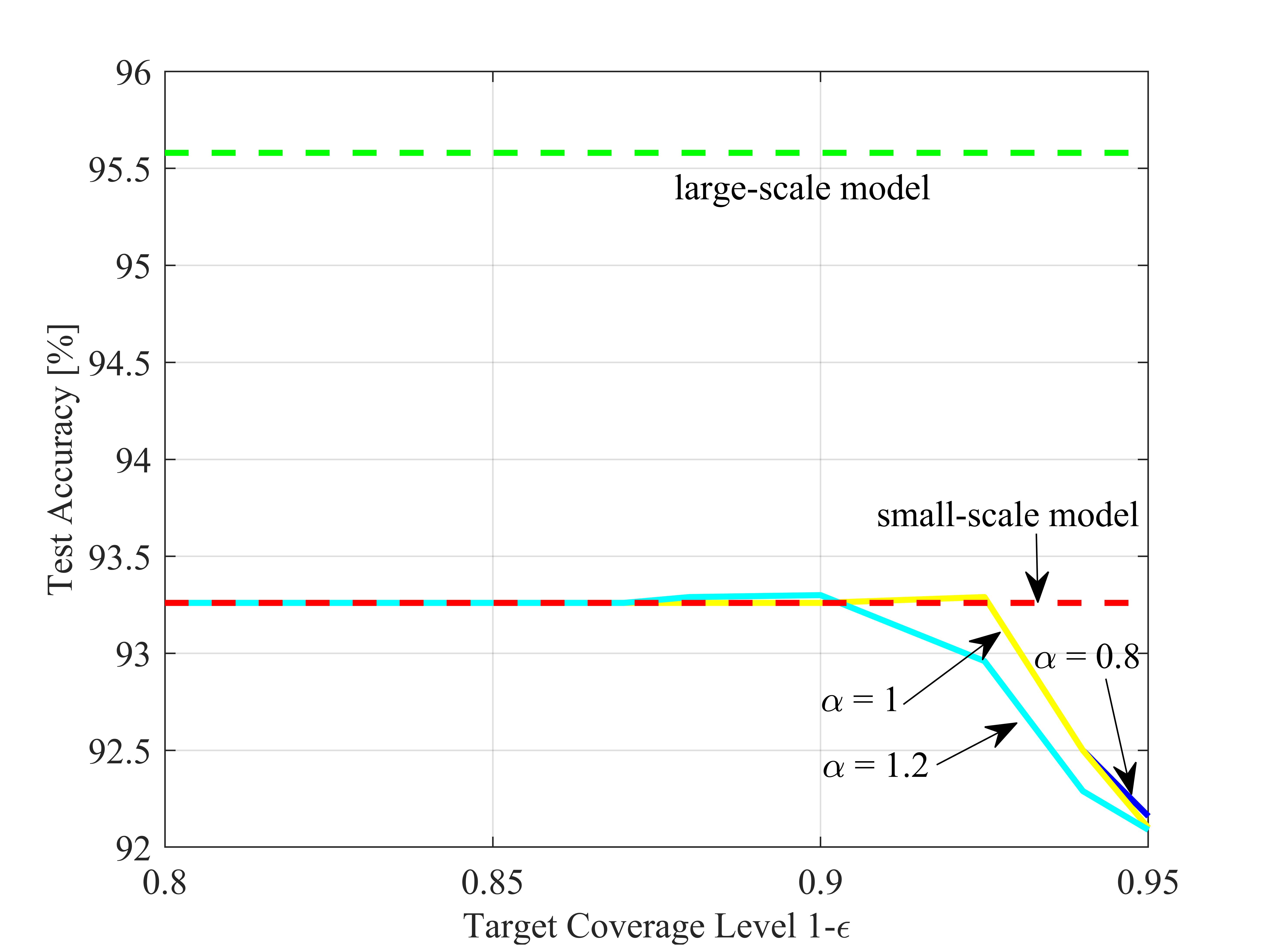}}
    \caption{Accuracy versus target coverage rate $1-\epsilon$ for different values of $\alpha$ for the $\alpha$-divergence used in (\ref{eq:credal_set}) on the CIFAR-10 data set with classes $\{\text{airplane, automobile, bird}\}$. The dashed lines report the accuracy performance of the large-scale model predictive distribution $p^*(\cdot|x)$, and of the small-scale model $p(\cdot|x)$. Note that there is no change to accuracy when applying Laplace approximation in a post-processing way.}
    \label{result:cifar_accuracy_versus_coverage} 
\end{figure}

While Fig.~\ref{result:cifar_changing_accuracy} focuses on the performance of the credal set $\Gamma(x)$, we now turn to analyzing the performance of the predictive distribution (\ref{eq:intersection_non_normalized}) extracted from the subset $\Gamma(x)$ as described in Sec.~\ref{sec:intersection_prob}. To this end, Fig.~\ref{result:cifar_ECE_versus_coverage} and Fig.~\ref{result:cifar_accuracy_versus_coverage} plot the ECE and accuracy of the small-scale model versus the target coverage level $1-\epsilon$ in the range from $0.8$ to $0.95$, respectively. We also vary the $\alpha$ values used in evaluating the subset (\ref{eq:credal_set}). These figures also report for reference the ECE and accuracy performance of the large-scale model $p^*(\cdot|x)$, and of the Laplace approximation method $q^{\text{La}}(\cdot|x)$ in (\ref{eq:laplace_prob}).

The key observation from Fig.~\ref{result:cifar_ECE_versus_coverage} and Fig.~\ref{result:cifar_accuracy_versus_coverage} is that, through a suitable choice of the hyperparameters $\epsilon$ and $\alpha$, CD-CI can improve the ECE of the original, uncalibrated, small-scale model, as well as of the Laplace approximation, without any accuracy loss. Moreover, the performance is robust to the choice of the hyperparameter $\epsilon$. In fact, with all values within the range $0.85 \leq 1-\epsilon \leq 0.92$, CD-CI outperforms both the original small-scale model and the Laplace approximation in terms of ECE, achieving ECE reductions of approximately $4\%$ and $3\%$, respectively. For smaller values of $1-\epsilon$, CD-CI tends to produce credal set $\Gamma(x)$ concentrated around the small-scale model predictive distribution. Conversely, for larger values of $1-\epsilon$, the credal set $\Gamma(x)$ becomes larger, eventually leading to a deterioration in the ECE and of the accuracy of CD-CI. 

\begin{figure} [htb] 
    \centering
    \centerline{\includegraphics[scale=0.045]{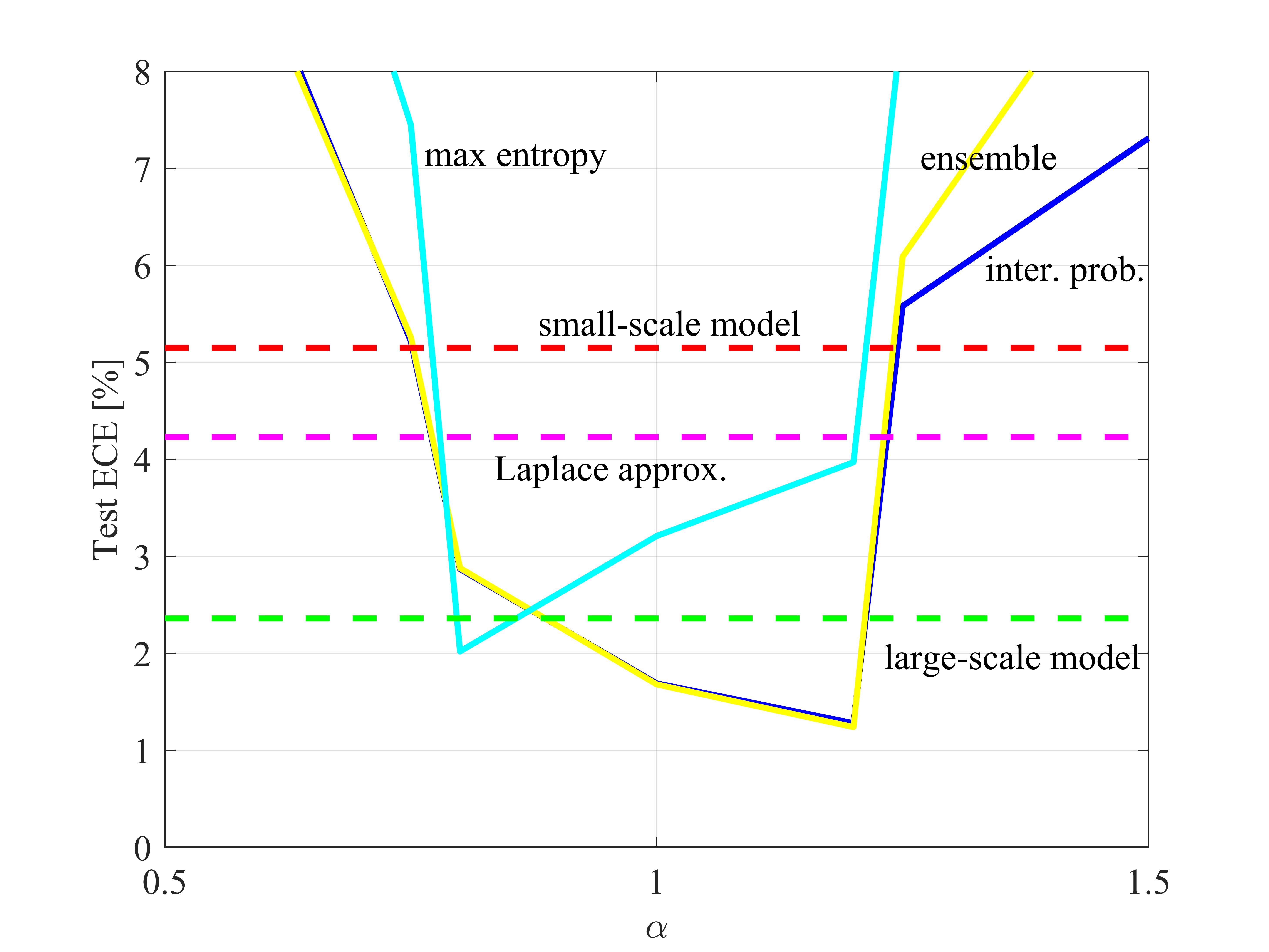}}
    \caption{ECE versus the values of $\alpha$ for the $\alpha$-divergence used in (\ref{eq:credal_set}) for different target coverage rate $1-\epsilon$ on the CIFAR-10 data set with classes $\{\text{airplane, automobile, bird}\}$. The blue line represents the ECE performance of the predictive distribution $ q^*(y|x)$ (\ref{eq:intersection_non_normalized}), in yellow the ECE of the ensemble distribution $\mathbb{E}_{\Gamma(x)}[q(\cdot|x)]$, and in cyan the ECE of the maximum Shannon entropy distribution $\max_{q(\cdot|x) \in \Gamma(x)}[H(q(\cdot|x)]$.}
    \label{result:cifar_ablation_fix_epsilon} 
\end{figure}

\begin{figure} [htb] 
    \centering
    \centerline{\includegraphics[scale=0.045]{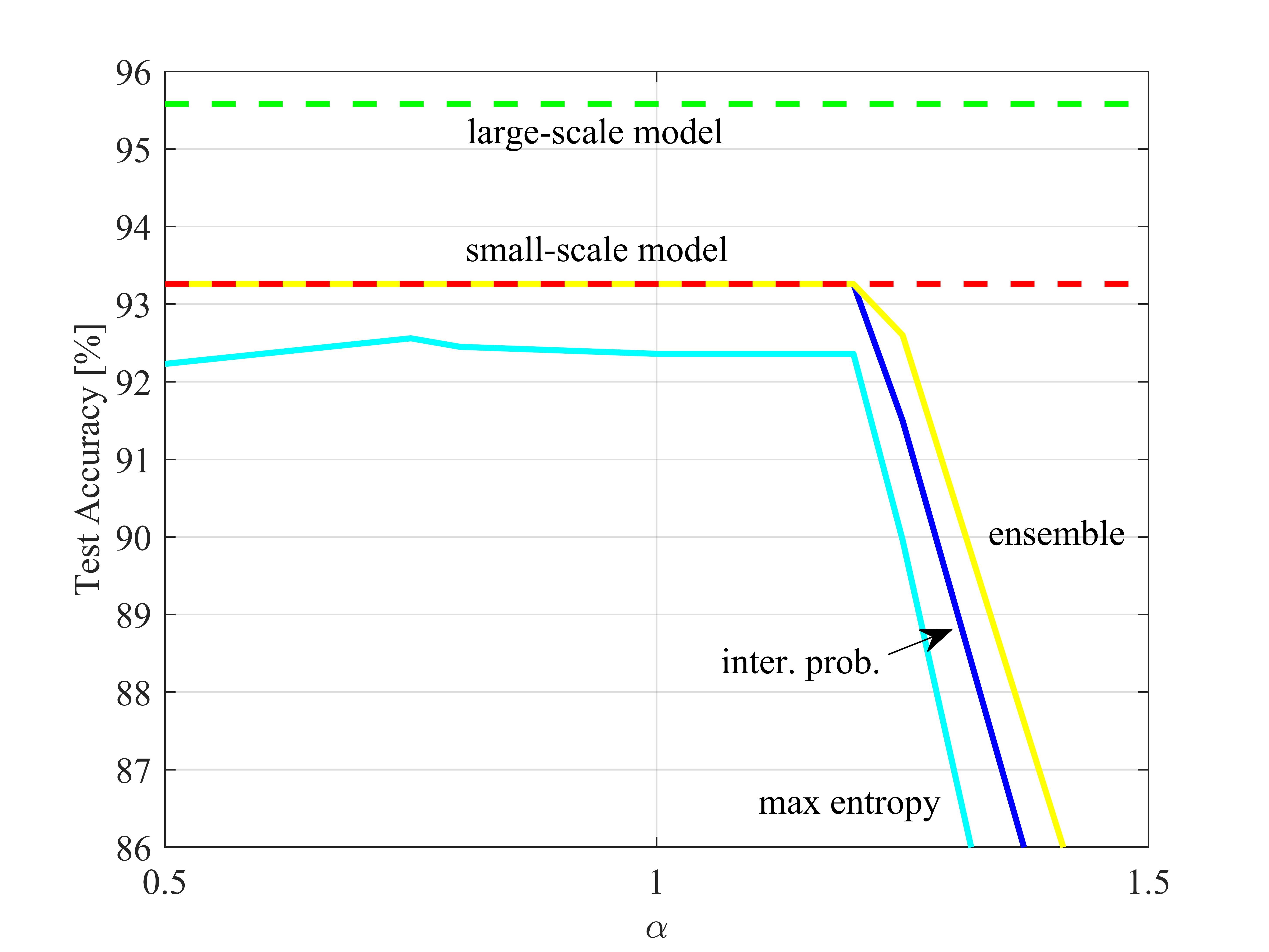}}
    \caption{Accuracy versus the values of $\alpha$ for the $\alpha$-divergence used in (\ref{eq:credal_set}) for different target coverage rate $1-\epsilon$ on the CIFAR-10 data set with classes $\{\text{airplane, automobile, bird}\}$. 
    Legend is as in Figure~\ref{result:cifar_ablation_fix_epsilon}. 
    }
    \label{result:cifar_ablation_fix_epsilon_accuracy} 
\end{figure}

Fig.~\ref{result:cifar_ECE_versus_coverage} and Fig.~\ref{result:cifar_accuracy_versus_coverage} also illustrate the advantages of constraining the credal set through the choice of a value of $\alpha$ larger than $\alpha=1$, thus moving beyond the KL divergence. To investigate the impact of the divergence parameter $\alpha$, the ECE and the accuracy are shown in Fig.~\ref{result:cifar_ablation_fix_epsilon} and Fig.~\ref{result:cifar_ablation_fix_epsilon_accuracy}, respectively, as a function of $\alpha$ for $1-\epsilon=0.9$. In these figures, we evaluate and compare three approaches for deriving a single predictive distribution, namely (\emph{i}) the intersection probability (\ref{eq:intersection_non_normalized}); (\emph{ii}) the ensemble distribution $\mathbb{E}_{\Gamma(x)}[q(\cdot|x)]$, where the average is over a uniform distribution in the credal set $\Gamma(x)$; and (\emph{iii}) the maximum Shannon entropy distribution $\max_{q(\cdot|x) \in \Gamma(x)}[H(q(\cdot|x)]$ \cite{caprio2024conformalized}.


The results confirm that the intersection probability approach in (\ref{eq:intersection_non_normalized}) consistently achieves the best performance in terms of ECE, outperforming both the ensemble and maximum Shannon entropy methods. Moreover, with the intersection probability method, the best results are obtained with $\alpha$ larger than 1, although excessively large values of $\alpha$ cause a performance degradation.


\subsection{Natural Language Classification}
For the natural language task, we adopt the SNLI data set \cite{bowman2015large}, an English language sentence pairs classification data set with three labels: entailment, contradiction, and neutral. We use the NLI-deberta-v3-large adopted temperature scaling \cite{shen2024thermometer} and original NLI-deberta-v3-small \cite{he2020deberta} as the large-scale and small-scale models, respectively. We further vary the quality of the small-scale model via uniform quantization.

\begin{figure} [htb] 
    \centering
    \centerline{\includegraphics[scale=0.16]{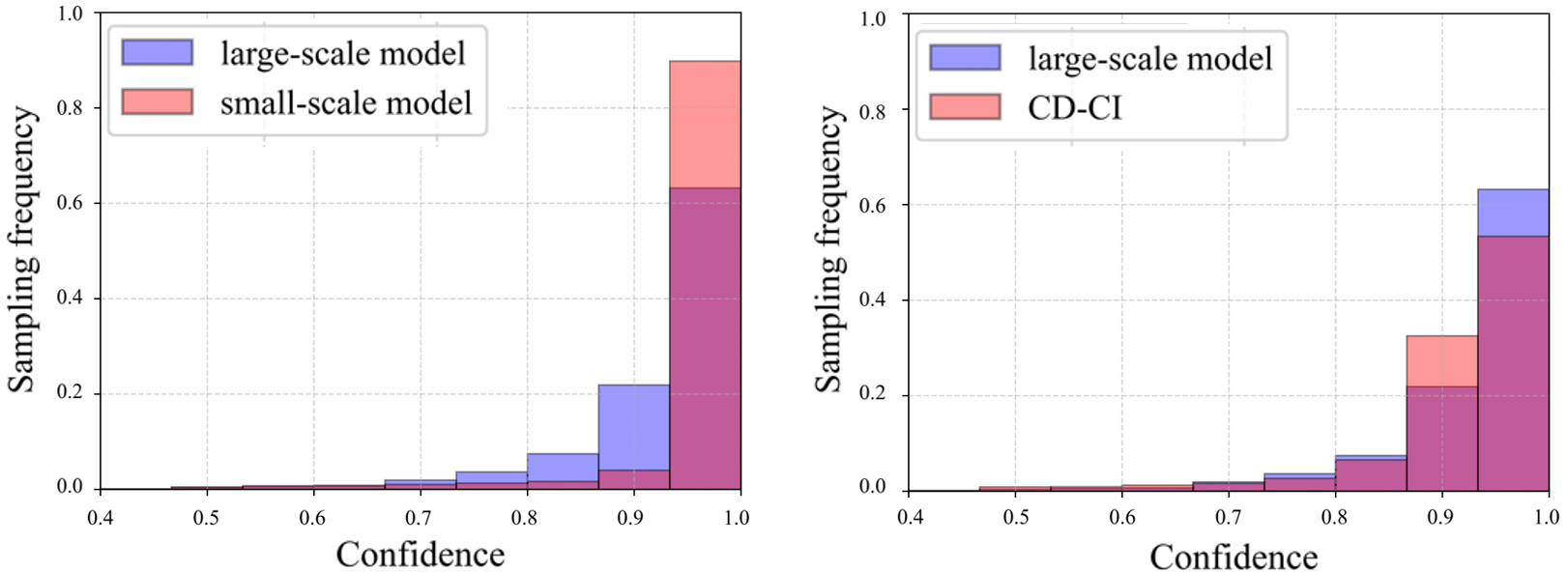}}
    \caption{Confidence histograms for the large-scale model predictive distribution $p^*(y|x)$, for the small-scale model $p(y|x)$, and for CD-CI $q^*(y|x)$ evaluated on SNLI data set, using $\alpha = 1$ with target coverage rate $1-\epsilon = 0.9$.}
    \label{result:snli_visualization} 
\end{figure}

Firstly, to visualize the calibration benefits of CD-CI, Fig.~\ref{result:snli_visualization} plots the confidence histograms evaluated over test data $(x,y)$ for the large-scale model predictive distribution $p^*(y|x)$, the small-scale model $p(y|x)$, and CD-CI $q^*(y|x)$ on the SNLI data set, using $\alpha = 1$ with target coverage rate $1-\epsilon = 0.9$. As illustrated, while the small-scale model is overconfident, CD-CI successfully recalibrate the model, ensuring decisions that are calibrated to a level similar to the reference large-scale model.

\begin{figure} [htb] 
    \centering
    \centerline{\includegraphics[scale=0.23]{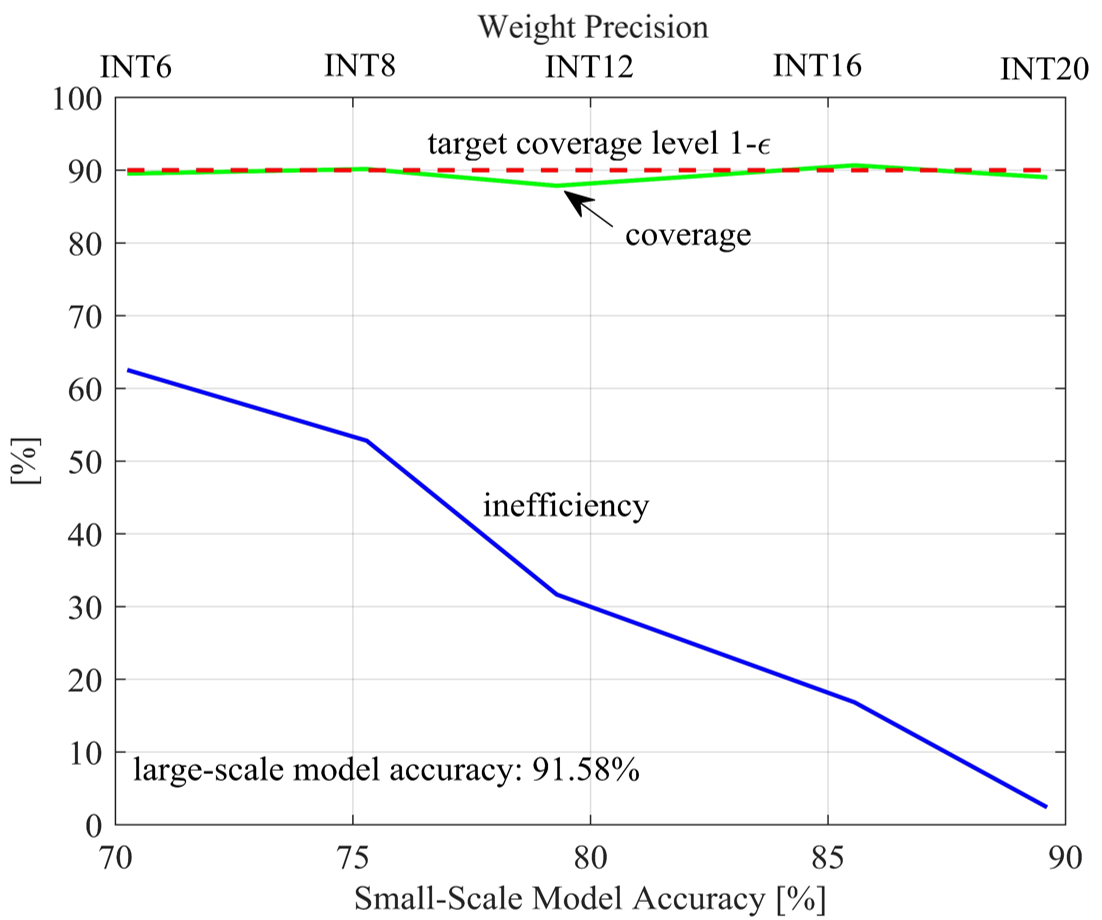}}
    \caption{Coverage and inefficiency versus the small-scale models accuracy on the SNLI data set using the KL divergence in (\ref{eq:credal_set}) with target coverage rate $1-\epsilon = 0.9$. The accuracy of the large-scale model is $91.58\%$. Note that INT$X$ denotes the block-floating point format, in which each element of the block is expressed via INT$X$ precision (see e.g., \cite{li2024evaluating, drumond2018training}).}
    \label{result:snli_changing_accuracy} 
\end{figure}

Then, with the same settings as in Fig.~\ref{result:cifar_changing_accuracy}, we further investigate the relationship between the accuracy of the small-scale model and the performance of the CD-CI in Fig.~\ref{result:snli_changing_accuracy}. To achieve this, we control the accuracy of the small-scale model through uniform quantization with varying weight precision \cite{li2024evaluating, leviathan2023fast}, a strategy that facilitates the efficient edge deployment of Transformer network architectures. The results in Fig.~\ref{result:snli_changing_accuracy} reaffirm the conclusions drawn in Fig.~\ref{result:cifar_changing_accuracy}.

\begin{figure} [htb] 
    \centering
    \centerline{\includegraphics[scale=0.21]{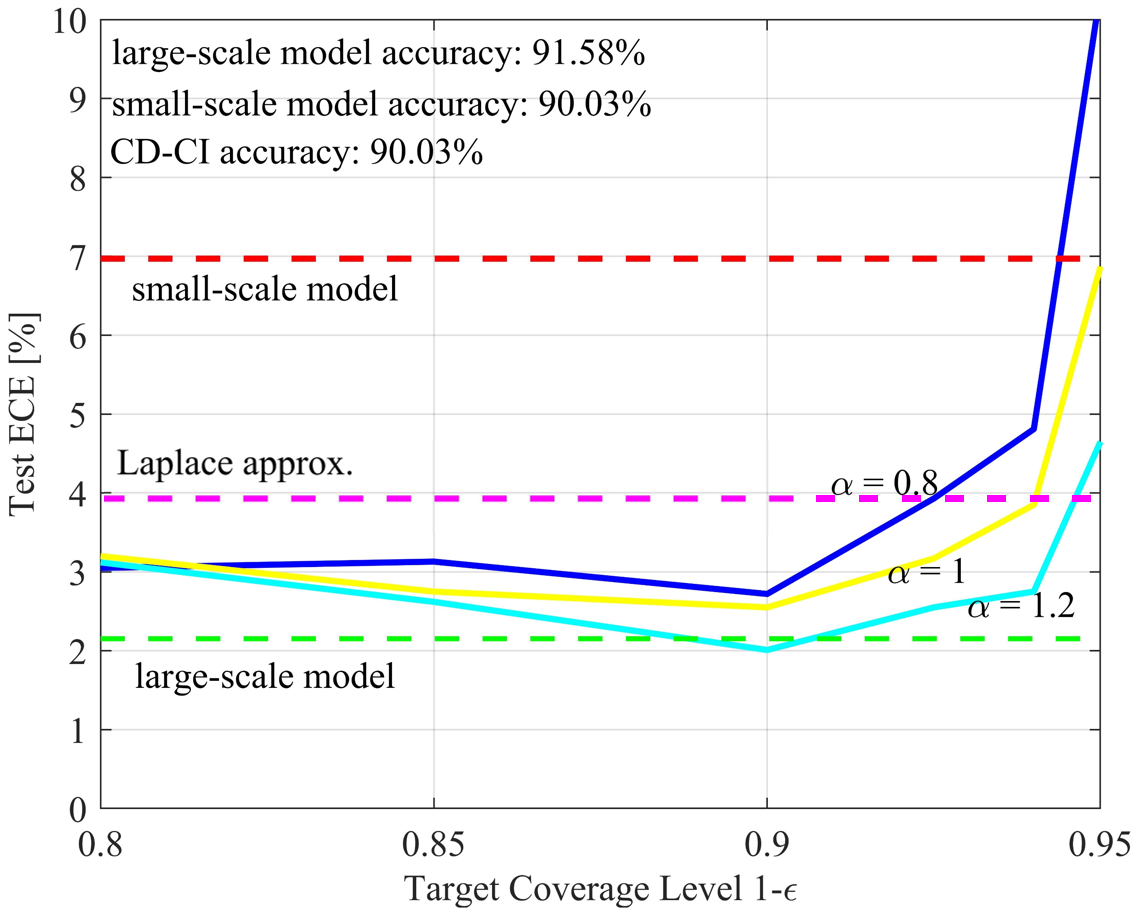}}
    \caption{ECE versus target coverage rate $1-\epsilon$ for different values of the $\alpha$ for $\alpha$-divergence used in (\ref{eq:credal_set}) on the SNLI data set. The dashed lines report the ECE performance of the large-scale model predictive distribution $p^*(\cdot|x)$, of the small-scale model $p(\cdot|x)$, and of the Laplace approximation method $q^{\text{La}}(\cdot|x)$ in (\ref{eq:laplace_prob}).}
    \label{result:snli_ECE_versus_coverage} 
\end{figure}

\begin{figure} [htb] 
    \centering
    \centerline{\includegraphics[scale=0.045]{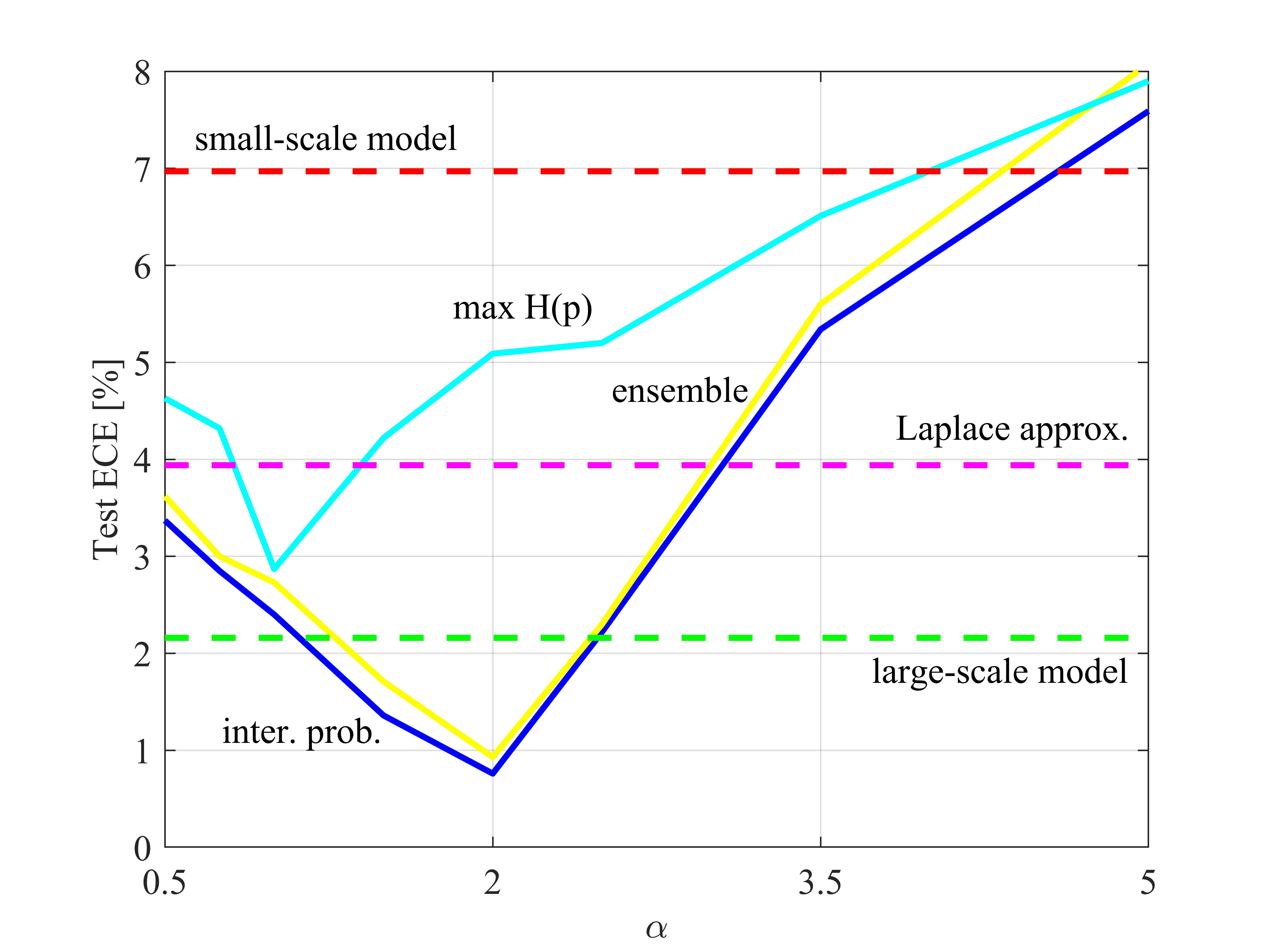}}
    \caption{ECE versus the values of $\alpha$ for the $\alpha$-divergence used in (\ref{eq:credal_set}) for different target coverage rate $1-\epsilon$ on the SNLI data set. The solid lines represent the ECE performance of the predictive distribution $ q^*(y|x)$ (\ref{eq:intersection_non_normalized}), of the ensemble distribution $\mathbb{E}_{\Gamma(x)}[q(\cdot|x)]$, and of the maximum Shannon entropy distribution $\max_{q(\cdot|x) \in \Gamma(x)}[H(q(\cdot|x)]$.}
    \label{result:snli_ablation_fix_epsilon} 
\end{figure}

\begin{figure} [htb] 
    \centering
    \centerline{\includegraphics[scale=0.045]{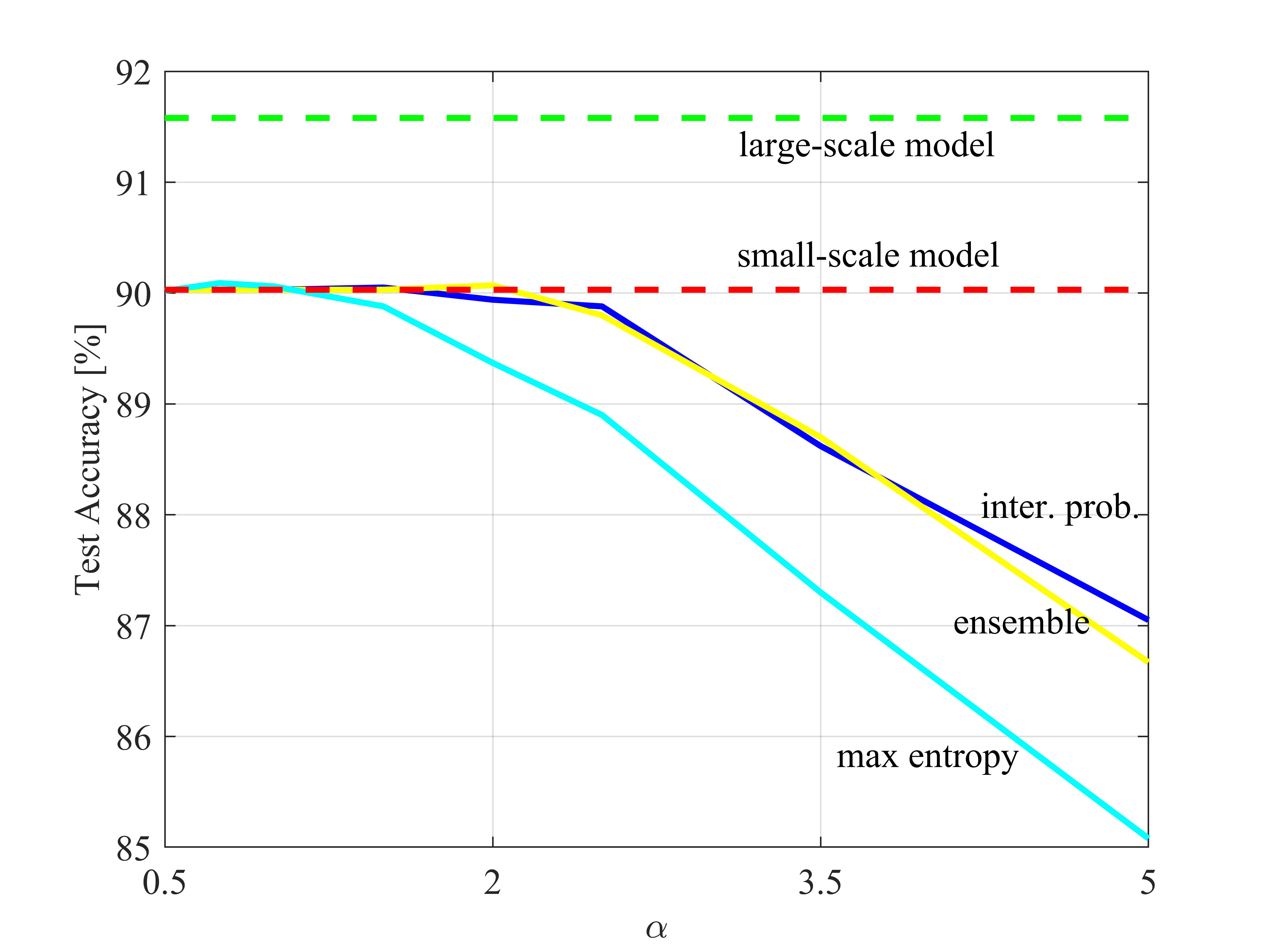}}
    \caption{Accuracy versus target coverage rate $1-\epsilon$ for $\alpha =1$ for the $\alpha$-divergence used in (\ref{eq:credal_set}) on the SNLI data set. The solid lines represent the accuracy of the predictive distribution $ q^*(y|x)$ (\ref{eq:intersection_non_normalized}), of the ensemble distribution $\mathbb{E}_{\Gamma(x)}[q(\cdot|x)]$, and of the maximum Shannon entropy distribution $\max_{q(\cdot|x) \in \Gamma(x)}[H(q(\cdot|x)]$.}
    \label{result:/snli_ablation_fix_epsilon_accuracy} 
\end{figure}

Fig.~\ref{result:snli_ECE_versus_coverage} shows the ECE of CD-CI as a function of the target coverage level $1-\epsilon$, ranging from $80\%$ to $95\%$, for different values of $\alpha$. These results confirm the general conclusions from Fig.~\ref{result:cifar_ECE_versus_coverage}. In fact, in this experiment, the benefits of CD-CI are seen to be even more robust to the choice of hyperparameters $1-\epsilon$ and $\alpha$ in terms of the ECE. Overall, CD-CI achieves an improvement of approximately $5\%$ and $2\%$ in ECE compared to the original small-scale model and the Laplace approximation.


With the same settings as in Fig.~\ref{result:cifar_ablation_fix_epsilon} and Fig.~\ref{result:cifar_ablation_fix_epsilon_accuracy}, we evaluate the ECE and accuracy against the divergence hyperparameter $\alpha$ in Fig.~\ref{result:snli_ablation_fix_epsilon} and Fig.~\ref{result:/snli_ablation_fix_epsilon_accuracy}, respectively. These figures confirm again that (\emph{i}) the intersection approach (\ref{eq:intersection_non_normalized}) achieves superior performance in terms of ECE and accuracy, and (\emph{ii}) relatively large values of $\alpha$ improve performance, whereas excessively large values of $\alpha$ result in performance degradation.

\section{Conclusion} \label{sec:conclusion}
In this paper, we have proposed a low-complexity methodology to calibrate a small-scale edge model prior to deployment by leveraging data generated by a large-scale cloud model. The method, called Conformalized Distillation for Credal Inference (CD-CI), ensures that the edge model can produce, at runtime, a set of predictive distributions guaranteed to include the large-scale model's predictive distribution with a pre-specified probability. Unlike standard Bayesian learning techniques, the ensemble of predictive distributions is obtained through a simple thresholding operation applied directly to the small model's output.  

Future research directions may include evaluating the performance of conformalized credal inference under covariate shift  \cite{tibshirani2019conformal}, integrating the approach with prior-data fitted networks \cite{muller2021transformers}, exploring online calibration strategies that enable interactive communication between edge and cloud models \cite{leviathan2023fast}, and studying the interplay between credal sets and imprecise highest density regions \cite{caprio2024conformalized}.

\bibliographystyle{IEEEtran}
\bibliography{refs}

\end{document}